\pdfoutput=1

\documentclass[11pt]{article}

\usepackage{naacl2021}
\usepackage{times}
\usepackage{latexsym}

\usepackage[T1]{fontenc}

\usepackage[utf8]{inputenc}

\usepackage{microtype}

\usepackage{url}

\usepackage{graphicx}
\usepackage{url}
\usepackage{booktabs}
\usepackage{float}

\usepackage{amsmath}
\usepackage{amssymb}
\usepackage{multirow}

\usepackage{xspace}

\usepackage[normalem]{ulem}

\usepackage{nccmath}

\title{Multilingual BERT Post-Pretraining Alignment}

\author{Lin Pan$\dagger$, Chung-Wei Hang$\dagger$, Haode Qi$\dagger$, Abhishek Shah$\dagger$, Saloni Potdar$\dagger$, Mo Yu$\ddagger$ \\
  $\dagger$IBM Watson \\
  $\ddagger$MIT-IBM Watson AI Lab \\
  {\small\tt \{panl,hangc\}@us.ibm.com, \{Haode.Qi, Abhishek.Shah1\}@ibm.com, \{potdars, yum\}@us.ibm.com} \\}

\date{}

\begin{document}
\maketitle

\begin{abstract}
We propose a simple method to align multilingual contextual embeddings as a post-pretraining step for improved cross-lingual transferability of the pretrained language models. Using parallel data, our method aligns embeddings on the word level through the recently proposed Translation Language Modeling objective as well as on the sentence level via contrastive learning and random input shuffling. We also perform sentence-level code-switching with English when finetuning on downstream tasks. On XNLI, our best model (initialized from mBERT) improves over mBERT by $4.7\%$ in the zero-shot setting and achieves comparable result to XLM for translate-train while using less than $18\%$ of the same parallel data and $31\%$ fewer model parameters. On MLQA, our model outperforms XLM-R$_{\text{Base}}$, which has $57\%$ more parameters than ours.
 

\end{abstract}

\section{Introduction}
\label{sec:introduction}




Building on the success of monolingual pretrained language models (LM) such as BERT \cite{Devlin+18} and RoBERTa \cite{Liu+19}, their multilingual counterparts mBERT \cite{Devlin+18} and XLM-R \cite{Conneau+20a} are trained using the same objectives---\textbf{Masked Language Modeling (MLM)} and in the case of mBERT, Next Sentence Prediction (NSP). MLM is applied to monolingual text that covers over $100$ languages. Despite the absence of parallel data and explicit alignment signals, these models transfer surprisingly well from high resource languages, such as English, to other languages. On the Natural Language Inference (NLI) task XNLI \cite{Conneau+18}, a text classification model trained on English training data can be directly applied to the other $14$ languages and achieve respectable performance. Having a single model that can serve over $100$ languages also has important business applications.


Recent work improves upon these pretrained models by adding cross-lingual tasks leveraging parallel data that always involve English. \citet{Conneau-Lample-19} pretrain a new Transformer-based \cite{Vaswani+17} model from scratch with an MLM objective on monolingual data, and a Translation Language Modeling (TLM) objective on parallel data.
\citet{Cao+20} align mBERT embeddings in a post-hoc manner:
They first apply a statistical toolkit, FastAlign \cite{Dyer+13}, to create word alignments on parallel sentences. Then, mBERT is tuned via minimizing the mean squared error between the embeddings of English words and those of the corresponding words in other languages.
Such post-hoc approach suffers from the limitations of word-alignment toolkits: (1) the noises from FastAlign can lead to error propagation to the rest of the pipeline; (2) FastAlign mainly creates the alignments with word-level translation and usually overlooks the contextual semantic compositions. As a result, the tuned mBERT is biased to shallow cross-lingual correspondence. Importantly, both approaches only involve word-level alignment tasks.

In this work, we focus on self-supervised, alignment-oriented training tasks using minimum parallel data to improve mBERT's cross-lingual transferability. We propose a \textbf{Post-Pretraining Alignment (PPA)} method consisting of both word-level and sentence-level alignment, as well as a finetuning technique on downstream tasks that take pairs of text as input, such as NLI and Question Answering (QA). Specifically, we use a slightly different version of TLM as our word-level alignment task and contrastive learning \cite{Hadsell+06} on mBERT's \texttt{[CLS]} tokens to align sentence-level representations. Both tasks are self-supervised and do not require pre-alignment tools such as FastAlign. Our sentence-level alignment is implemented using MoCo \cite{He+20}, an instance discrimination-based method of contrastive learning that was recently proposed for self-supervised representation learning in computer vision. Lastly, when finetuning on NLI and QA tasks for non-English languages, we perform sentence-level code-switching with English as a form of both alignment and data augmentation. We conduct controlled experiments on XNLI and MLQA \cite{Lewis+20}, leveraging varying amounts of parallel data during alignment. We then conduct an ablation study that shows the effectiveness of our method. On XNLI, our aligned mBERT improves over the original mBERT by $4.7\%$ for zero-shot transfer, and outperforms \citet{Cao+20} while using the same amount of parallel data from the same source. For translate-train, where translation of English training data is available in the target language, our model achieves comparable performance to XLM while using far fewer resources. On MLQA, we get $2.3\%$ improvement over mBERT and outperform XLM-R$_{\text{Base}}$ for zero-shot transfer.


\section{Method}
\label{sec:approach}
\begin{figure}
    \centering
    \includegraphics[width=.4\textwidth]{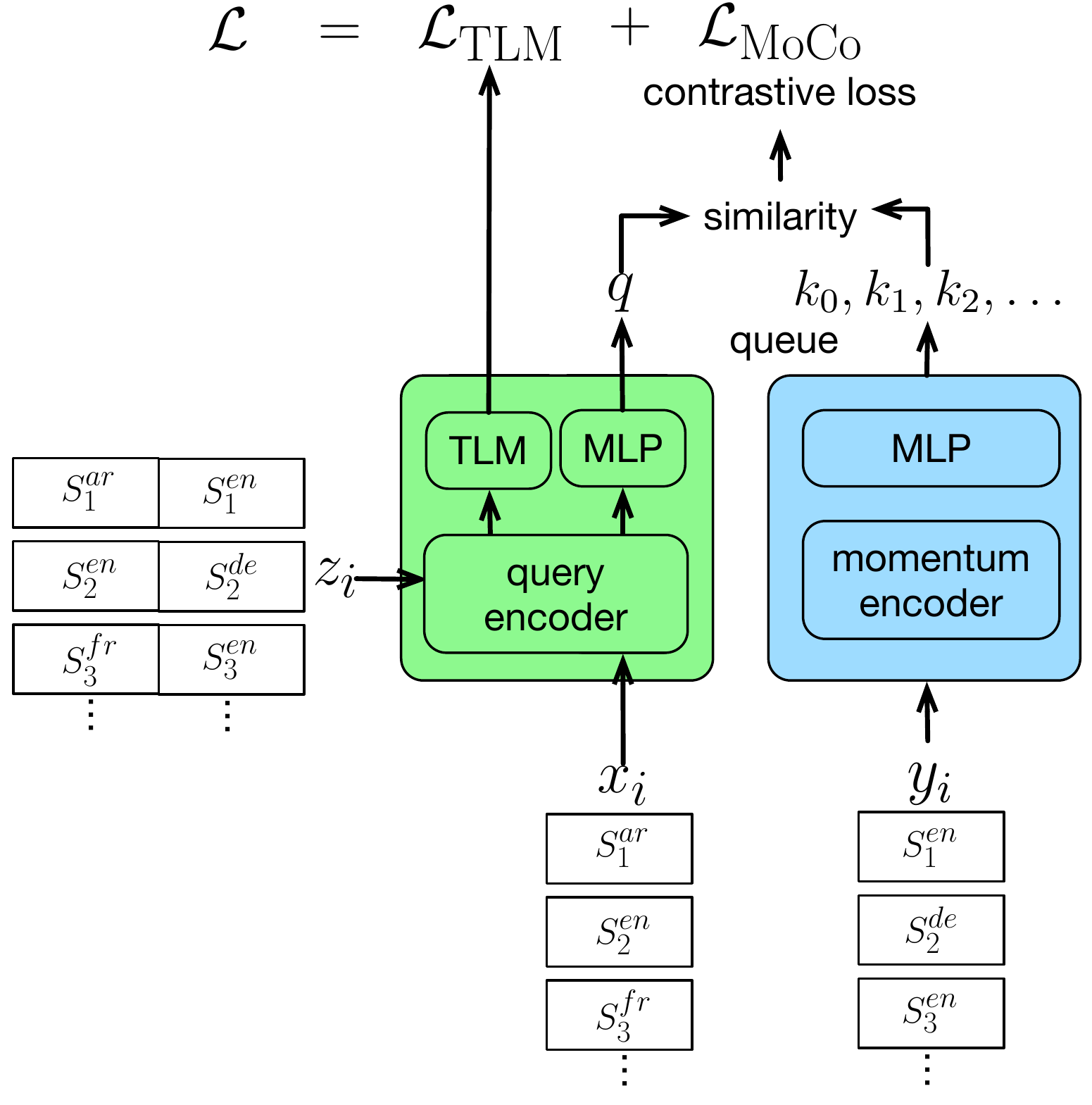}\vspace{-2mm}
    \caption{Model structure for our Post-Pretraining Alignment method using parallel data. We use MoCo to implement our sentence-level objective and TLM for our word-level objective. The model is trained in a multi-task manner with both objectives.}
    \label{fig:approach}
\end{figure}

This section introduces our proposed Post-Pretraining Alignment (PPA) method. We first describe the MoCo contrastive learning framework and how we use it for sentence-level alignment. Next, we describe the finer-grained word-level alignment with TLM. Finally, when training data in the target language is available, we incorporate sentence-level code-switching as a form of both alignment and data augmentation to complement PPA.
%
Figure~\ref{fig:approach} shows our overall model structure.


%
\paragraph{Background: Contrastive Learning}
Instance discrimination-based contrastive learning aims to bring two views of the same source image closer to each other in the representation space while encouraging views of different source images to be dissimilar through a contrastive loss. Recent advances in this area, such as SimCLR \cite{Chen+2020} and MoCo \cite{He+20} have bridged the gap in performance between self-supervised representation learning and fully-supervised methods on the ImageNet \cite{Deng+09} dataset. As a key feature for both methods, a large number of negative examples per instance are necessary for the models to learn such good representations. SimCLR uses in-batch negative example sampling, thus requiring a large batch size, whereas MoCo stores negative examples in a queue and casts the contrastive learning task as dictionary (query-key) lookup. In what follows, we first describe MoCo and then how we use it for sentence-level alignment. 

Concretely, MoCo employs a dual-encoder architecture. Given two views $v_1$ and $v_2$ of the same image, $v_1$ is encoded by the query encoder $f_q$ and $v_2$ by the momentum encoder $f_k$. $v_1$ and $v_2$ form a positive pair. Negative examples are views of different source images, and are stored in a queue $\in{K}$, which is randomly initialized. $K$ is usually a large number (e.g., $K=65,536$ for ImageNet). Negative pairs are formed by comparing $v_1$ with each item in the queue. Similarity between pairs is measured by dot product. MoCo uses the InfoNCE loss \cite{Oord+19} to bring positive pairs closer to each other and push negative pairs apart. After a batch of view pairs are processed, those encoded by the momentum encoder are added to the queue as negative examples for future queries. During training, the query encoder is updated by the optimizer while the momentum encoder is updated by the exponential moving average of the query encoder's parameters to maintain queue consistency:
\begin{equation}\label{eq1}
\theta_k = m\theta_k + (1 - m)\theta_q
\end{equation}
where $\theta_q$ and $\theta_k$ are model parameters of $f_q$ and $f_k$, respectively. $m$ is the momentum coefficient. 

\subsection{Sentence-Level Alignment Objective}
Our sentence-level alignment falls under the general problem of bringing two views of inputs from the same source closer in the representation space while keeping those from different sources \mbox{dissimilar} through a contrastive loss.
From a cross-lingual alignment perspective, we treat an English sequence $S_{i}^{en}$ and its translation $S_{i}^{tr}$ in another language $tr \in L$ as two manifestations of the same semantics. At the same time, sentences that are not translations of each other should be further apart in the representation space. Given parallel corpora consisting of $\{(S_{1}^{en}, S_{1}^{tr}),\ldots, (S_{N}^{en}, S_{N}^{tr})\}$, we align sentence representations in all the different languages together using MoCo.

We use the pretrained mBERT model to initialize both the query and momentum encoders. mBERT is made of $12$ Transformer blocks, $12$ attention heads, and hidden size $d_h = 768$. For input, instead of feeding the query encoder with English examples and the momentum encoder with translation examples or vice versa, we propose a \textbf{random input shuffling} approach. Specifically, we randomly shuffle the order of $S_{i}^{en}$ and $S_{i}^{tr}$ when feeding the two encoders, so that the query encoder sees both English and translation examples. We observe that this is a crucial step towards learning good multilingual representations using our method. The final hidden state $h\in \mathbb{R}^{1\times d_h}$ of the \texttt{[CLS]} token, normalized with $L_2$ norm, is treated as the sentence representation \footnote{Alternatively, we also experimented with mean-pooling of the last layer's embeddings as the sentence representation, but it performed slightly worse than using the \texttt{[CLS]} token.}. Following \citet{Chen+2020}, we add a non-linear projection layer on top of $h$:
\begin{equation}
z = W_2 ReLU (W_1 h),\label{eq2}
\end{equation}
where $W_1\in \mathbb{R}^{d_h\times d_h}$, $W_2\in \mathbb{R}^{d_k\times d_h}$, and $d_k$ is set to $300$. The model is trained using the InfoNCE loss:
\begin{equation}
\mathcal{L_{\text{MoCo}}} = -\log \frac{\exp(z_q \cdot z_{k+} /\tau)}{\sum_{k=1}^K \exp(z_q \cdot z_k /\tau)},\label{eq3}
\end{equation}
where $\tau$ is a temperature parameter. In our implementation, we use a relatively small batch size of $128$, resulting in more frequent parameter updates than if a large batch size were used. Items enqueued early on can thus become outdated with a large queue, so we scale down the queue size to $K=32,000$ to prevent the queue from becoming stale.

\subsection{Word-Level Alignment Objective}
We use TLM for word-level alignment. TLM is an extension of MLM that operates on bilingual data---parallel sentences are concatenated and MLM is applied to the combined bilingual sequence. Different from \citet{Conneau-Lample-19}, we do not reset positional embeddings when forming the bilingual sequence, and we also do not use language embeddings. In addition, the order of $S_{i}^{en}$ and $S_{i}^{tr}$ during concatenation is determined by the random input shuffling from the sentence-level alignment step and we add a \texttt{[SEP]} token between $S_{i}^{en}$ and $S_{i}^{tr}$. 

We randomly mask $15\%$ of the WordPiece tokens in each combined sequence. Masking is done by using a special \texttt{[MASK]} token $80\%$ of the times, a random token in the vocabulary $10\%$ of the times, and unchanged for the remaining $10\%$. TLM is performed using the query encoder of MoCo. Our final PPA model is trained in a multi-task manner with both sentence-level objective and TLM:
\begin{equation}\label{eq4}
\mathcal{L} = \mathcal{L_{\text{MoCo}}} + \mathcal{L_{\text{TLM}}},
\end{equation}

\subsection{Finetuning on Downstream Tasks}
After an alignment model is trained with PPA, we extract the query encoder from MoCo and finetune it on downstream tasks for evaluation. We follow the standard way of finetuning BERT-like models for sequence classification and QA tasks: (1) on XNLI, we concatenate the premise with the hypothesis, and add a \texttt{[SEP]} token in between. A softmax classifier is added on top of the final hidden state of the \texttt{[CLS]} token;
(2) on MLQA, we concatenate the question with the context, and add a \texttt{[SEP]} token in between. We add two linear layers on top of mBERT followed by softmax over the context tokens to predict answer start and end positions, respectively. 

We conduct experiments in two settings: 1. \emph{Zero-shot cross-lingual transfer}, where training data is available in English but not in target languages. 2. \emph{Translate-train}, where the English training set is (machine) translated to all the target languages.  For the latter setting, we perform data augmentation with code-switched inputs, when training on languages other than English. For example, a Spanish question $q_{es}$ and context $c_{es}$ pair can be augmented to \emph{two} question-context pairs ($q_{es}$, $c_{en}$) and ($q_{en}$, $c_{es}$) with code-switching, resulting in $2$x training data \footnote{The original question-context pair ($q_{es}$, $c_{es}$) is not used for training as it did not help improve model performance in our experiments.}. The same goes for XNLI with premises and hypotheses. The code-switching is always between English, and a target language. During training, we ensure the two augmented pairs appear in the same batch.

\begin{table*}[htb]
    \begin{minipage}{\textwidth}
    \centering
    \scalebox{0.7}{
    \begin{tabular}{l|ccccccc|r}
    \toprule
    {\bf Resource} & \bf{fr} & \bf{es} & \bf{de} & \bf{bg} & \bf{ar} & \bf{zh} & \bf{hi} & \bf{total} \\
    \midrule
    \multicolumn{8}{l}{\em Original data} \\
    \midrule
    MultiUN    & 14.2M & 12.2M &     - &     - & 10.6M & 10.5M &    - & \\
    Europarl   &  2.1M &  2.0M &  2.0M &  0.4M &     - &     - &    - & \\
    EUbookshop &     - &     - &  9.6M &  0.2M &     - &     - &    - & \\
    IITB       &     - &     - &     - &     - &     - &     - & 1.6M & \\
    \midrule
    \multicolumn{8}{l}{\em Considered in this paper} \\
    \midrule
    MultiUN    &     - &     - &     - &     - & 10.6M & 10.5M &    - & \\
    Europarl   &  2.1M &  2.0M &  2.0M &  0.4M &     - &     - &    - & \\
    EUbookshop &     - &     - &     - &  0.2M &     - &     - &    - & \\
    IITB       &     - &     - &     - &     - &     - &     - & 1.6M & \\
    \em{Total} &  2.1M &  2.0M &  2.0M &  0.6M & 10.6M & 10.5M & 1.6M & \\
    \midrule
    \multicolumn{8}{l}{\em Used for our post-pretraining alignment (PPA)} \\
    \midrule
    {\bf Ours} (250k) & 250k & 250k & 250k & 250k & 250k & 250k & 250k &  1.8M \\
    {\bf Ours} (600k) & 600k & 600k & 600k & 467k & 600k & 600k & 600k &  4.1M \\
    {\bf Ours} (2M)   & 1.8M & 1.7M & 1.7M & 467k & 2.0M & 2.0M & 0.8M & 10.5M \\
    \midrule
    \multicolumn{8}{l}{\em Used by other approaches} \\
    \midrule
    \citet{Cao+20}\footnote{\citet{Cao+20} uses the same 250k parallel corpora as our 250k setting, thus giving an apple-to-apple comparison.} & 250k & 250k & 250k & 250k & 250k & 250k & 250k &  1.8M \\
    \citet{Artetxe-Schwenk-19}\footnote{\citet{Artetxe-Schwenk-19}'s number includes a total of $93$ languages.} & - & - & - & - & - & - & - & 223M \\
    XLM \cite{Conneau-Lample-19}\footnote{We only list the number of parallel sentences XLM uses for the languages we consider.} & 14.2M & 12.2M & 9.6M & 0.2M & 10.6M & 10.5M & 1.6M & 58.9M \\
    \bottomrule
    \end{tabular}
    }
    \caption{Parallel data statistics.  All parallel data involve English as source language.  We use Europarl for en-fr, en-es, and en-de, both Europarl and EUbookshop for en-bg, MultiUN for en-ar, en-zh, and IITB for en-hi.  Our 250k setting uses an equal amount of data from the same source as \citet{Cao+20}.  Our 2M setting uses approximately 63\% and 17.8\% of the parallel data \citet{Artetxe-Schwenk-19} and \citet{Conneau-Lample-19} use, respectively.}
    \label{tab:parallel-data-stats}
    \end{minipage}
\end{table*}

\section{Experimental Settings}
\label{sec:settings}

\subsection{Parallel Data for Post-Pretraining}

\paragraph{Parallel Data}
All parallel data we use involve English as the source language. Specifically, we collect en-fr, en-es, en-de parallel pairs from \mbox{Europarl}, en-ar, en-zh from MultiUN 
\cite{Ziemski+16}, en-hi from IITB \cite{Kunchukuttan+18}, and en-bg from both Europarl and \mbox{EUbookshop}.  All datasets were downloaded from the OPUS\footnote{\url{http://opus.nlpl.eu/}} website \cite{Tiedemann-12}.  In our experiments, we vary the number of parallel sentence pairs for PPA.  For each language, we take the first 250k, 600k, and 2M English-translation parallel sentence pairs except for those too short (where either sentence has less than $10$ WordPiece tokens), or too long (where both sentences concatenated together have more than $128$ WordPiece tokens).  Table~\ref{tab:parallel-data-stats} shows the actual number of parallel pairs in each of our 250k, 600k, and 2M settings.

\subsection{Evaluation Benchmarks}

\paragraph{XNLI} is an evaluation dataset for cross-lingual NLI that covers $15$ languages. The dataset is human-translated from the development and test sets of the English MultiNLI dataset \cite{Williams+18}. Given a sentence pair of premise and hypothesis, the task is to classify their relationship as \emph{entailment}, \emph{contradiction}, and \emph{neutral}. For zero-shot cross-lingual transfer, we train on the English MultiNLI training set, and apply the model to the test sets of the other languages. For translate-train, we train on translation data that come with the dataset \footnote{\url{https://cims.nyu.edu/~sbowman/xnli/}}.

\paragraph{MLQA} is an evaluation dataset for QA that covers seven languages. The dataset is derived from a three step process. (1) Parallel sentence \mbox{mining} from Wikipedia of the languages. (2) English question annotation and answer span annotation on English context. (3) Professional translation of English questions to the other languages as well as answer span annotation. MLQA has two evaluation tasks: (a) Cross-lingual transfer (XLT), where the question and context are in the same language. (b) Generalized cross-lingual transfer (G-XLT), where the question and context are in different languages. We focus on XLT in this work. For zero-shot cross-lingual transfer, we train on the English SQuAD v1.1 \cite{Rajpurkar+16} training set. For translate-train, we train on translation data provided in \citet{Hu+20} \footnote{\url{https://github.com/google-research/xtreme}}

\subsection{Training Details}
For both PPA and finetuning on downstream tasks, we use the AdamW optimizer with $0.01$ weight decay and a linear learning rate scheduler. For PPA, we use a batch size of $128$, mBERT max sequence length $128$ and learning rate warmup for the first $10\%$ of the total iterations, peaking at $0.00003$. The MoCo momentum is set to $0.999$, queue size $32000$ and temperature $0.05$. Our PPA models are trained for $10$ epochs, except for the 2M setting where $5$ epochs are trained. On XNLI, we use a batch size of $32$, mBERT max sequence length $128$ and finetune the PPA model for $2$ epochs. Learning rate peaks at $0.00005$ and warmup is done to the first $1000$ iterations. On MLQA, mBERT max sequence length is set to $386$ and peak learning rate $0.00003$. The other parameters are the same as XNLI. Our experiments are run on a single $32$ GB V100 GPU, except for PPA training that involves either MLM or TLM, where two such GPUs are used. We also use mixed-precision training to save on GPU memory and speed up experiments.
\section{Results}
\label{sec:evaluation}

\label{sec:datasets}


We report results on the test set of XNLI and MLQA and we do hyperparameter searching on the development set. All the experiments for translate-train were done using the code-switching technique introduced in Section~\ref{sec:approach}. 

\begin{table*}[htb]
    \centering
    \scalebox{0.8}{
    \begin{tabular}{l|cccccccc|c}
    \toprule
    {\bf Model} & \bf{en} & \bf{fr} & \bf{es} & \bf{de} & \bf{bg} & \bf{ar} & \bf{zh} & \bf{hi} & \bf{avg} \\
    \midrule
    \multicolumn{10}{c}{\emph{Zero-shot cross-lingual transfer}} \\
    \midrule
    mBERT \cite{Devlin+18}       & 81.4 &    - & 74.3 & 70.5  &    - & 62.1 & 63.8 &  - & - \\
    mBERT from \cite{Hu+20}      & 80.8 & 73.4 & 73.5 & 70.0  & 68.0 & 64.3 & 67.8 & 58.9 & 69.6 \\
    \citet{Cao+20}                & 80.1 & 74.5 & 75.5 & 73.1  & 73.4 &    - &    - &    - & - \\
    \citet{Artetxe-Schwenk-19}    & 73.9 & 71.9 & 72.9 & 72.6  & 74.2 & \bf{71.4} & 71.4 & 65.5 & 71.7 \\ 
    {\bf Ours} (250k)            & 82.4 & 75.5 & 76.2 & 73.3  & \bf{74.6} & 68.2 & 71.7 & 62.8 & 73.1 \\
    {\bf Ours} (600k)            & 82.4 & \bf{76.7} & 76.4 & 74.0  & 74.1 & 69.1 & 72.3 & 66.9 & 74.0 \\
    {\bf Ours} (2M)              & \bf{82.8} & 76.6 & \bf{76.7} & \bf{74.2}  & 73.8 & 70.3 & \bf{72.8} & \bf{66.9} & \bf{74.3} \\
    XLM (MLM)                    & 83.2 & 76.5 & 76.3 & 74.2  & 74.0 & 68.5 & 71.9 & 65.7 & 73.8 \\
    XLM (MLM + TLM)              & \underline{85.0} & \underline{78.7} & \underline{78.9} & \underline{77.8}  & \underline{77.4} & \underline{73.1} & \underline{76.5} & \underline{69.6} & \underline{77.1} \\
    \midrule
    \multicolumn{10}{c}{\emph{Translate-train}} \\
    \midrule
    mBERT \cite{Devlin+18}            & 81.9 &    - & 77.8 & 75.9  &    - & 70.7 & 76.6 &     - & - \\
    mBERT from \cite{Wu-Dredze-19}    & 82.1 & 76.9 & 78.5 & 74.8  & 75.4 & 70.8 & 76.2 & 65.3 & 75.0 \\ 
    {\bf Ours} (250k)                 & 82.4 & 78.8 & 79.0 & \bf{78.7}  & 78.4 & 74.0 & 77.9 & 69.6 & 77.4 \\
    {\bf Ours} (600k)                 & 82.4 & 79.7 & 79.7 & 77.9  & \bf{79.0} & 75.2 & 77.8 & 71.5 & 77.9 \\
    {\bf Ours} (2M)                   & \bf{82.8} & \bf{79.7} & \bf{80.6} & 78.6  & 78.8 & \bf{75.2} & \bf{78.0} & \bf{72.0} & \bf{78.2} \\
    XLM \cite{Conneau-Lample-19}      & \underline{85.0} & \underline{80.2} & \underline{80.8} & \underline{80.3}  & \underline{79.3} & \underline{76.5} & \underline{78.6} & \underline{72.3} & \underline{79.1} \\
    \bottomrule
    \end{tabular}%
    }
    \caption{XNLI accuracy scores for each language. After alignment, our best model improves over mBERT by $4.7\%$ for zero-shot transfer, and achieves comparable performance to XLM for translate-train. \citet{Artetxe-Schwenk-19} use 223M parallel sentences covering $93$ languages. XLM uses 58.9M parallel sentences (for the seven languages we consider) with 40\% more parameters.  Our approach (250k, 600k, and 2M per language) uses a total of 1.8M, 4.1M, and 10.5M parallel sentences, respectively.}
    \label{tab:main-results}
\end{table*}
\paragraph{XNLI}
Table~\ref{tab:main-results} shows results on XNLI measured by accuracy. \citet{Devlin+18} only provide results on a few languages\footnote{\url{https://github.com/google-research/bert/blob/master/multilingual.md}}, so we use the mBERT results from \citet{Hu+20} as our baseline for zero-shot cross-lingual transfer, and \citet{Wu-Dredze-19} for translate-train. Our best model, trained with 2M parallel sentences per language improves over mBERT baseline by $4.7\%$ for zero-shot transfer, and $3.2\%$ for translate-train. 

Compared to \citet{Cao+20}, which use 250k parallel sentences per language from the same sources as we do for post-pretraining alignment, our 250k model does better for all languages considered and we do not rely on the word-to-word pre-alignment step using FastAlign, which is prone to error propagation to the rest of the pipeline.

Compared to XLM, our 250k, 600k and 2M settings represent $3.1\%$, $7\%$ and $17.8\%$ of the parallel data used by XLM, respectively (see Table~\ref{tab:parallel-data-stats}). The XLM model also has $45\%$ more parameters than ours as Table~\ref{tab:model-parameters} shows. Furthermore, XLM trained with MLM only is already significantly better than mBERT even though the source of its training data is the same as mBERT from Wikipedia. One reason could be that XLM contains $45\%$ more model parameters than mBERT as model depth and capacity are shown to be key to cross-lingual success \cite{K+20}. Additionally, \citet{Wu-Dredze-19} hypothesize that limiting pretraining to the languages used by downstream tasks may be beneficial since XLM models are pretrained on the $15$ XNLI languages only. Our 2M model bridges the gap between mBERT and XLM from $7.5\%$ to $2.8\%$ for zero-shot transfer. Note that, for bg, our total processed pool of en-bg data consists of 456k parallel sentences, so there is no difference in en-bg data between our 600k and 2M settings. For translate-train, our model achieves comparable performance to XLM with the further help of code-switching during finetuning. 
\begin{table}[htb]
    \centering
    \scalebox{0.7}{
    \begin{tabular}{l|rrrrrrr}
    \toprule
    \bf Model & \bf \# langs & \bf L & $\mathbf{H_m}$ & \bf $\mathbf{H_{ff}}$ & \bf A & \bf V & \bf \# params \\
    \midrule
    mBERT                 & 104 & 12 &  768 & 3072 & 12 & 110k & 172M \\ 
    XLM                   &  15 & 12 & 1024 & 4096 &  8 &  95k & 250M \\
    XLM-R$_{\text{Base}}$ & 100 & 12 &  768 & 3072 & 12 & 250k & 270M \\
    {\bf Ours}            & 104 & 12 &  768 & 3072 & 12 & 110k & 172M \\ 
    \bottomrule
    \end{tabular}%
    }
    \caption{Model architecture and sizes from \citet{Conneau+20a}.  $L$ is the number of Transformer layers, $H_m$ is the hidden size, $H_{ff}$ is the dimension of the feed-forward layer, $A$ is the number of attention heads, and $V$ is the vocabulary size.}
    \label{tab:model-parameters}
\end{table}

Our alignment-oriented method is, to a large degree, upper-bounded by the English performance, since all our parallel data involve English and all the other languages are implicitly aligning with English through our PPA objectives. Our 2M model is able to improve the English performance to $82.4$ from the mBERT baseline, but it is still lower than XLM (MLM), and much lower than XLM (MLM+TLM). We hypothesize that more high-quality monolingual data and model capacity are needed to further improve our English performance, thereby helping other languages better align with it.

\paragraph{MLQA}
Table~\ref{tab:main-results-mlqa} shows results on MLQA measured by F1 score. We notice the mBERT baseline from the original MLQA paper is significantly lower than that from \citet{Hu+20}, so we use the latter as our baseline. Our 2M model outperforms the baseline by $2.3\%$ for zero-shot and is also $0.2\%$ better than XLM-R$_{\text{Base}}$, which uses $57\%$ more model parameters than mBERT as Table~\ref{tab:model-parameters} shows. For translate-train, our 250k model is $1.3\%$ better than the baseline. 

Comparing our model performance using varying amounts of parallel data, we observe that 600k per language is our sweet spot considering the trade-off between resource and performance. Going up to 2M helps on XNLI, but less significantly compared to the gain going from 250k to 600k. On MLQA, surprisingly, 250k slightly outperforms the other two for translate-train.  
\begin{table*}[htb]
    \centering
    \scalebox{0.8}{
    \begin{tabular}{l|cccccc|c}
    \toprule
    {\bf Model} & \bf{en} & \bf{ar} & \bf{de} & \bf{es} & \bf{hi} & \bf{zh} & \bf{avg} \\
    \midrule
    \multicolumn{7}{c}{\emph{Zero-shot cross-lingual transfer}} \\
    \midrule
    mBERT from \cite{Lewis+20}       & 77.7 & 45.7 & 57.9 & 64.3 & 43.8 & 57.5 & 57.8 \\
    mBERT from \cite{Hu+20}          & \bf{80.2} & 52.3 & 59.0 & 67.4 & 50.2 & 59.6 & 61.5 \\
    {\bf Ours} (250k)           & 80.0 & 52.6 & \bf{63.2} & 67.7 & 54.1 & 60.5 & 63.0 \\
    {\bf Ours} (600k)           & 79.7 & 52.4 & 62.8 & 67.6 & 58.3 & 60.4 & 63.5 \\
    {\bf Ours} (2M)             & 79.8 & 53.8 & 62.3 & 67.7 & 57.9 & 61.5 & \bf{63.8} \\
    XLM from \cite{Lewis+20}         & 74.9 & 54.8 & 62.2 & \bf{68.0} & 48.8 & 61.1 & 61.6 \\
    XLM-R$_{\text{Base}}$ \cite{Conneau+20a} & 77.1 & \bf{54.9} & 60.9 & 67.4 & \bf{59.4} & \bf{61.8} & 63.6 \\
    \midrule
    \multicolumn{7}{c}{\emph{Translate-train}} \\
    \midrule
    mBERT from \cite{Lewis+20} & 77.7 & 51.8 & 62.0 & 53.9 & 55.0 & 61.4 & 60.3 \\
    mBERT from \cite{Hu+20}    & \bf{80.2} & 55.0 & 64.6 & 70.0 & 60.1 & 63.9 & 65.6 \\ 
    {\bf Ours} (250k)          & 80.0 & 58.0 & \bf{65.7} & \bf{71.0} & 62.0 & \bf{64.4} & \bf{66.9} \\
    {\bf Ours} (600k)          & 79.7 & 58.1 & 65.2 & 70.5 & \bf{63.4} & 64.1 & 66.8 \\
    {\bf Ours} (2M)            & 79.8 & \bf{58.2} & 64.7 & 70.6 & 63.1 & 64.4 & 66.8 \\
    XLM from \cite{Lewis+20}   & 74.9 & 54.0 & 61.4 & 65.2 & 50.7 & 59.8 & 61.0 \\
    \bottomrule
    \end{tabular}}
    \caption{MLQA F1 scores for each language. After alignment, our best model improves over mBERT baseline by $2.3\%$ and outperforms XLM-R$_{\text{Base}}$ for zero-shot transfer. Our model trained with the smallest amount of parallel data is $1.3\%$ better than mBERT baseline for translate-train.}
    \label{tab:main-results-mlqa}
\end{table*}

\paragraph{Ablation}
\begin{table*}[htb]
    \centering
    \scalebox{0.7}{
    \begin{tabular}{l|cccccccc|c}
    \toprule
    {\bf Model} & \bf en & \bf fr & \bf es & \bf de & \bf bg & \bf ar & \bf zh & \bf hi & \bf avg \\
    \midrule
    \multicolumn{10}{c}{\emph{Zero-shot cross-lingual transfer}} \\
    \midrule
    Our full system (250k) & \bf{82.4} & \bf{75.5} & \bf{76.2} & \bf{73.3} & \bf{74.6} & \bf{68.2} & \bf{71.7} & 62.8 & \bf{73.1} \\
    \quad - MoCo & 82.2 & 75.3 & 75.8 & 73.0 & 71.3 & 67.1 & 71.3 & 61.8 & 72.2 \\
    \quad - TLM & 80.5 & 74.7 & 75.2 & 71.4 & 72.7 & 66.2 & 68.9 & \bf{64.0} & 71.7 \\
    \quad repl TLM w/ MLM & 81.5 & 75.0 & 75.2 & 70.8 & 72.5 & 66.2 & 69.0 & 61.9 & 71.5 \\
    \midrule
    Our full system (600k) & \bf{82.4} & \bf{76.7} & \bf{76.4} & \bf{74.0} & \bf{74.1} & \bf{69.1} & \bf{72.3} & \bf{66.9} & \bf{74.0} \\
    \quad - MoCo & 82.0 & 75.5 & 75.9 & 72.8 & 72.1 & 68.5 &72.1 & 64.5 & 72.9 \\
    \quad - TLM     & 81.2 & 75.1 & 75.4 & 71.9 & 73.3 & 68.2 & 71.0 & 65.8 & 72.7 \\
    \quad repl TLM w/ MLM  & 82.2 & 75.7 & 75.5 & 73.0 & 73.3 & 68.5 & 71.1 & 66.5 & 73.2 \\
    \midrule
    Our full system (2M)   & \bf{82.8} & \bf{76.6} & \bf{76.7} & \bf{74.2} & \bf{73.8} & \bf{70.3} & \bf{72.8} & \bf{66.9} & \bf{74.3} \\
    \quad - MoCo & 82.5 & 75.2 & 76.3 & 72.4 & 71.9 & 67.9 & 71.4 & 65.2 & 72.9 \\
    \quad - TLM  & 81.3 & 76.2 & 76.4 & 73.2 & 72.9 & 69.0 & 71.5 & 66.1 & 73.3 \\
    \quad repl TLM w/ MLM   & 82.0 & 75.8 & 75.8 & 73.2 & 73.5 & 68.7 & 70.6 & 65.8 & 73.2 \\
    \midrule
    \multicolumn{10}{c}{\emph{Translate-train}} \\
    \midrule
    Our full system (250k)        & \bf{82.4} & 78.8 & 79.0 & \bf{78.7} & 78.4 & \bf{74.0} & \bf{77.9} & 69.6 & \bf{77.4} \\
    \quad - MoCo           & 82.2 & \bf{79.8} & \bf{79.8} & 77.8 & \bf{78.9} & 73.8 & 77.3 & 69.8 & 77.4 \\
    \quad - TLM            & 80.5 & 78.3 & 77.8 & 77.5 & 77.4 & 72.4 & 77.2 & 69.5 & 76.3 \\
    \quad repl TLM w/ MLM        & 81.5 & 78.4 & 79.4 & 78.3 & 78.2 & 73.4 & 76.9 & \bf{69.9} & 77.0 \\
    \quad - CS & 82.4 & 77.8 & 79.5 & 76.2 & 76.2 & 73.2 & 77.5 & 67.9 & 76.3 \\
    \midrule
    Our full system (600k) & \bf{82.4} & \bf{79.7} & \bf{79.7} & 77.9 & \bf{79.0} & \bf{75.2} & 77.8 & \bf{71.5} & \bf{77.9} \\
    \quad - MoCo   & 82.0 & 79.5 & 79.2 & \bf{78.1} & 78.9 & 74.1 & \bf{78.1} & 71.0 & 77.6 \\
    \quad - TLM     & 81.2 & 78.5 & 78.6 & 78.1 & 77.7 & 73.7 & 76.6 & 70.8 & 76.9 \\
    \quad repl TLM w/ MLM & 82.2 & 78.4 & 78.4 & 77.1 & 78.0 & 73.9 & 76.9 & 70.8 & 77.0 \\
    \quad - CS & 82.4 & 79.2 & 78.3 & 77.5 & 77.0 & 73.6 & 77.3 & 69.9 & 76.9 \\
    \midrule
    Our full system (2M)        & \bf{82.8} & \bf{79.7} & \bf{80.6} & 78.6 & \bf{78.8} & 75.2 & \bf{78.0} & \bf{72.0} & \bf{78.2} \\
    \quad - MoCo           & 82.5 & 79.1 & 80.0 & \bf{79.1} & 78.5 & \bf{75.3} & 77.7 & 70.5 & 77.8 \\ 
    \quad - TLM            & 81.3 & 78.9 & 79.4 & 78.0 & 77.8 & 74.4 & 77.2 & 70.0 & 77.1 \\
    \quad repl TLM w/ MLM        & 82.0 & 79.1 & 79.0 & 78.2 & 77.8 & 74.3 & 77.7 & 70.4 & 77.3 \\
    \quad - CS & 82.8 & 79.1 & 79.0 & 78.0 & 77.5 & 73.6 & 77.1 & 69.5 & 77.1 \\
    \bottomrule
    \end{tabular}}
    \caption{Ablation Study on XNLI.  250k, 600k, 2M refer to the maximum number of parallel sentence pairs per language used in PPA. \emph{MoCo} refers to our sentence-level alignment task using contrastive learning. \emph{TLM} refers to our word-level alignment task with translation language modeling. \emph{CS} stands for code-switching. We conduct an additional study \emph{repl TLM w/ MLM}, which means instead of TLM training, we augment our sentence-level alignment with regular MLM on monolingual text. This ablation confirms that the TLM objective helps because of its word alignment capability, not because we train the encoders with more data and iterations.}
    \label{tab:ablation-results}
\end{table*}
Table~\ref{tab:ablation-results} shows the contribution of each component of our method on XNLI. 
Removing TLM (\emph{-TLM}) consistently leads to about 1\% accuracy drop across the board, showing positive effects of the word-alignment objective.
To better understand TLM's consistent improvement, we replace TLM with MLM (\emph{repl TLM w/ MLM}), where we treat $S_{i}^{en}$ and $S_{i}^{tr}$ from the parallel corpora as separate monolingual sequences and perform MLM on each of them. The masking scheme is the same as TLM described in Section~\ref{sec:approach}. We observe that MLM does not bring significant improvement. This confirms that the improvement of TLM is not from the encoders being trained with more data and iterations. Instead, the word-alignment nature of TLM does help the multilingual training.

Comparing our model without word-level alignment, i.e., \emph{-TLM}, to the baseline mBERT in Table~\ref{tab:main-results}, we get 2--4\% improvement in the zero-shot setting and 1--2\% improvement in translate-train as the amount of parallel data is increased. These are relatively large improvements considering the fact that only sentence-level alignment is used. This also conforms to our intuition that sentence-level alignment is a good fit here since XNLI is a sentence-level task.

In the zero-shot setting, removing MoCo (\mbox{\emph{-MoCo}}) performs similarly to \emph{-TLM}, where we observe an accuracy drop of about 1\% compared to our full system. In translate-train, \emph{-MoCo} outperforms \emph{-TLM} and even matches the full system performance for 250k.

Finally, we show ablation result for our code-switching in translate-train. On average, code-switching provides an additional gain of $1\%$. 

\section{Related Work}

\paragraph{Training Multilingual LMs with Shared Vocabulary}
mBERT \cite{Devlin+18} is trained using MLM and NSP objectives on Wikipedia data in 104 languages with a shared vocabulary. Several works study what makes this pretrained model multilingual, and why it works well for cross-lingual transfer. \citet{Pires+20} hypothesize that having a shared vocabulary for all languages helps mapping tokens to a shared space. However, \citet{K+20} train several bilingual BERT models such as en-es, and {\em enfake}-es, where data for {\em enfake} is constructed by Unicode shifting of the English data such that there is no character overlap with data of the other language. Result shows that {\em enfake}-es still transfers well to Spanish and the contribution from shared vocabulary is very small. The authors point out that model depth and capacity instead are the key factors contributing to mBERT's cross-lingual transferability. XLM-R \cite{Conneau+20a} improves over mBERT by training longer with more data from CommonCrawl, and without the NSP objective. In terms of model size, \mbox{XLM-R} uses over 3x more parameters than mBERT. Its base version, XLM-R$_{\text{Base}}$, is more comparable to mBERT with the same hidden size and number of attention heads, but a larger shared vocabulary.

\paragraph{Training Multilingual LMs with Parallel Sentences}
In addition to MLM on monolingual data, XLM \cite{Conneau-Lample-19} further improves their cross-lingual LM pretraining by introducing a new TLM objective on parallel data.  TLM concatenates source and target sentences together, and predicts randomly masked tokens.  Our work uses a slightly different version of TLM together with a contrastive objective to post-pretrain mBERT. Unlike XLM, our TLM does not reset positions of target sentences, and does not use language embeddings. We also randomly shuffle the order of source and target sentences. Another difference between XLM and our work is XLM has $45\%$ more parameters and uses more training data.  Similar to XLM, Unicoder \cite{Huang+19} pretrains LMs on multilingual corpora. In addition to MLM and TLM, they introduce three additional cross-lingual pretraining tasks: word recover, paraphrase classification, and mask language model. \citet{Yang+20} propose Alternating Language Modeling (ALM). On a pair of bilingual sequences, instead of TLM, they perform phrase-level code-switching and MLM on the code-switched sequence. ALM is pretrained on both monolingual Wikipedia data and 1.5B code-switched sentences. 

\paragraph{Training mBERT with Word Alignments}
\citet{Cao+20} post-align mBERT embeddings by first generating word alignments on parallel sentences that involve English. For each aligned word pair, the $L_2$ distance between their embeddings is minimized to train the model. In order to maintain original transferability to downstream tasks, a regularization term is added to prevent the target language embeddings from deviating too much from their mBERT initialization. Our approach post-aligns mBERT with two self-supervised signals from parallel data without using pre-alignment tools.  \citet{Wang+19} also align mBERT embeddings using parallel data.  They learn a linear transformation that maps a word embedding in a target language to the embedding of the aligned word in the source language.  They show that their transformed embeddings are more effective on zero-shot cross-lingual dependency parsing.

Besides the aforementioned three major directions, \citet{Artetxe-Schwenk-19} train a multilingual sentence encoder on $93$ languages.  Their stacked BiLSTM encoder is trained by first generating embedding of a source sentence and then decoding the embedding into the target sentence in other languages.

Concurrent to our work, \citet{Chi+20}, \citet{Feng+20} and \citet{Wei+20} also leverage variants of contrastive learning for cross-lingual alignment. We focus on a smaller model and improve on it using as little parallel data as possible. We also explore code-switching during finetuning on downtream tasks to complement the \mbox{post-pretraining} alignment objectives.  


%
%

\section{Conclusion}
Post-pretraining embedding alignment is an efficient means of improving cross-lingual transferability of pretrained multilingual LMs, especially when pretraining from scratch is not feasible. We showed that our self-supervised sentence-level and word-level alignment tasks can greatly improve mBERT's performance on downstream tasks of NLI and QA, and the method can potentially be applied to improve other pretrained multilingual LMs. 

In addition to zero-shot cross-lingual transfer, we also showed that code-switching with English during finetuning provides additional alignment signals, when training data is available for the target language.

\bibliography{chang}

\begin{thebibliography}{30}
\expandafter\ifx\csname natexlab\endcsname\relax\def\natexlab#1{#1}\fi

\bibitem[{Artetxe and Schwenk(2019)}]{Artetxe-Schwenk-19}
Mikel Artetxe and Holger Schwenk. 2019.
\newblock Massively multilingual sentence embeddings for zero-shot
  cross-lingual transfer and beyond.
\newblock \emph{Transactions of the Association for Computational Linguistics
  (TACL)}, 7:597--610.

\bibitem[{Cao et~al.(2020)Cao, Kitaev, and Klein}]{Cao+20}
Steven Cao, Nikita Kitaev, and Dan Klein. 2020.
\newblock Multilingual alignment of contextual word representations.
\newblock In \emph{Proceedings of the 8th International Conference on Learning
  Representation (ICLR)}, Addis Ababa, Ethiopia.

\bibitem[{Chen et~al.(2020)Chen, Kornblith, Norouzi, and Hinton}]{Chen+2020}
Ting Chen, Simon Kornblith, Mohammad Norouzi, and Geoffrey Hinton. 2020.
\newblock \href {https://arxiv.org/abs/2002.05709} {A simple framework for
  contrastive learning of visual representations}.
\newblock \emph{arXiv preprint arXiv:2002.05709}, pages 1--18.

\bibitem[{Chi et~al.(2020)Chi, Dong, Wei, Yang, Singhal, Wang, Song, Mao,
  Huang, and Zhou}]{Chi+20}
Zewen Chi, Li~Dong, Furu Wei, Nan Yang, Saksham Singhal, Wenhui Wang, Xia Song,
  Xian-Ling Mao, Heyan Huang, and Ming Zhou. 2020.
\newblock \href {https://arxiv.org/abs/2007.07834} {{InfoXLM}: An
  information-theoretic framework for cross-lingual language model
  pre-training}.
\newblock \emph{arXiv preprint arXiv:2007.07834}, pages 1--11.

\bibitem[{Conneau et~al.(2020)Conneau, Khandelwal, Goyal, Chaudhary, Wenzek,
  Guzm\'{a}n, Grave, Ott, Zettlemoyer, and Stoyanov}]{Conneau+20a}
Alexis Conneau, Kartikay Khandelwal, Naman Goyal, Vishrav Chaudhary, Guillaume
  Wenzek, Francisco Guzm\'{a}n, Edouard Grave, Myle Ott, Luke Zettlemoyer, and
  Veselin Stoyanov. 2020.
\newblock Unsupervised cross-lingual representation learning at scale.
\newblock In \emph{Proceedings of the 58th Annual Meeting of the Association
  for Computational Linguistics (ACL)}, pages 8440--8451, Seattle.

\bibitem[{Conneau and Lample(2019)}]{Conneau-Lample-19}
Alexis Conneau and Guillaume Lample. 2019.
\newblock Cross-lingual language model pretraining.
\newblock In \emph{Advances in Neural Information Processing Systems (NIPS)},
  pages 7059--7069, Vancouver.

\bibitem[{Conneau et~al.(2018)Conneau, Lample, Rinott, Williams, Bowman,
  Schwenk, and Stoyanov}]{Conneau+18}
Alexis Conneau, Guillaume Lample, Ruty Rinott, Adina Williams, Samuel~R.
  Bowman, Holger Schwenk, and Veselin Stoyanov. 2018.
\newblock {XNLI}: Evaluating cross-lingual sentence representations.
\newblock In \emph{Proceedings of the Conference on Empirical Methods in
  Natural Language Processing (EMNLP)}, pages 2475--2485, Brussels.

\bibitem[{Deng et~al.(2009)Deng, Dong, Socher, Li, Li, and Fei-Fei}]{Deng+09}
Jia Deng, Wei Dong, Richard Socher, Li-Jia Li, Kai Li, and Li~Fei-Fei. 2009.
\newblock Image{N}et: A large-scale hierarchical image database.
\newblock In \emph{Processings of the {IEEE} Conference on Computer Vision and
  Pattern Recognition (CVPR)}, pages 248--255, Miami.

\bibitem[{Devlin et~al.(2019)Devlin, Chang, Lee, and Toutanova}]{Devlin+18}
Jacob Devlin, Ming-Wei Chang, Kenton Lee, and Kristina Toutanova. 2019.
\newblock {BERT}: Pre-training of deep bidirectional transformers for language
  understanding.
\newblock In \emph{Proceedings of the 20th Annual Conference of the North
  American Chapter of the Association for Computational Linguistics: Human
  Language Technologies (NAACL-HLT)}, pages 4171--4186, Minneapolis. The
  Association for Computational Linguistics.

\bibitem[{Dyer et~al.(2013)Dyer, Chahuneau, and Smith}]{Dyer+13}
Chris Dyer, Victor Chahuneau, and Noah~A. Smith. 2013.
\newblock A simple, fast, and effective reparameterization of {IBM} model 2.
\newblock In \emph{Proceedings of the Annual Conference of the North American
  Chapter of the Association for Computational Linguistics: Human Language
  Technologies (NAACL-HLT)}, pages 644--648, Atlanta. The Association for
  Computational Linguistics.

\bibitem[{Feng et~al.(2020)Feng, Yang, Cer, Arivazhagan, and Wang}]{Feng+20}
Fangxiaoyu Feng, Yinfei Yang, Daniel Cer, Naveen Arivazhagan, and Wei Wang.
  2020.
\newblock \href {https://arxiv.org/abs/2007.01852} {Language-agnostic {BERT}
  sentence embedding}.
\newblock \emph{arXiv preprint arXiv:2007.01852}, pages 1--13.

\bibitem[{Hadsell et~al.(2006)Hadsell, Chopra, and LeCun}]{Hadsell+06}
Raia Hadsell, Sumit Chopra, and Yann LeCun. 2006.
\newblock Dimensionality reduction by learning an invariant mapping.
\newblock In \emph{Proceedings of the {IEEE} Conference on Computer Vision and
  Pattern Recognition (CVPR)}, pages 1735--1742, New York.

\bibitem[{He et~al.(2020)He, Fan, Wu, Xie, and Girshick}]{He+20}
Kaiming He, Haoqi Fan, Yuxin Wu, Saining Xie, and Ross~B. Girshick. 2020.
\newblock Momentum contrast for unsupervised visual representation learning.
\newblock In \emph{Proceedings of the {IEEE/CVF} Conference on Computer Vision
  and Pattern Recognition (CVPR)}, pages 9726--9735, Seattle.

\bibitem[{Hu et~al.(2020)Hu, Ruder, Siddhant, Neubig, Firat, and
  Johnson}]{Hu+20}
Junjie Hu, Sebastian Ruder, Aditya Siddhant, Graham Neubig, Orhan Firat, and
  Melvin Johnson. 2020.
\newblock \href {https://arxiv.org/abs/2003.11080} {{XTREME}: A massively
  multilingual multi-task benchmark for evaluating cross-lingual
  generalization}.
\newblock \emph{arXiv preprint arXiv:2003.11080}, pages 1--20.

\bibitem[{Huang et~al.(2019)Huang, Liang, Duan, Gong, Shou, Jiang, and
  Zhou}]{Huang+19}
Haoyang Huang, Yaobo Liang, Nan Duan, Ming Gong, Linjun Shou, Daxin Jiang, and
  Ming Zhou. 2019.
\newblock Unicoder: A universal language encoder by pre-training with multiple
  cross-lingual tasks.
\newblock In \emph{Proceedings of the Conference on Empirical Methods in
  Natural Language Processing and the International Joint Conference on Natural
  Language Processing (EMNLP-IJCNLP)}, pages 2485--2494, Hong Kong.

\bibitem[{K et~al.(2020)K, Wang, Mayhew, and Roth}]{K+20}
Karthikeyan K, Zihan Wang, Stephen Mayhew, and Dan Roth. 2020.
\newblock Cross-lingual ability of multilingual {BERT}: An empirical study.
\newblock In \emph{Proceedings of the 8th International Conference on Learning
  Representation (ICLR)}, Addis Ababa, Ethiopia.

\bibitem[{Kunchukuttan et~al.(2018)Kunchukuttan, Mehta, and
  Bhattacharyya}]{Kunchukuttan+18}
Anoop Kunchukuttan, Pratik Mehta, and Pushpak Bhattacharyya. 2018.
\newblock The {IIT} {B}ombay {E}nglish-{H}indi parallel corpus.
\newblock In \emph{Proceedings of the Eleventh International Conference on
  Language Resources and Evaluation (LREC)}, pages 3473--3476, Miyazaki, Japan.

\bibitem[{Lewis et~al.(2020)Lewis, O\u{g}uz, Rinott, Riedel, and
  Schwenk}]{Lewis+20}
Patrick Lewis, Barlas O\u{g}uz, Ruty Rinott, Sebastian Riedel, and Holger
  Schwenk. 2020.
\newblock {MLQA}: Evaluating cross-lingual extractive question answering.
\newblock In \emph{Proceedings of the 58th Annual Meeting of the Association
  for Computational Linguistics (ACL)}, pages 1--16, Seattle.

\bibitem[{Liu et~al.(2019)Liu, Ott, Goyal, Du, Joshi, Chen, Levy, Lewis,
  Zettlemoyer, and Stoyanov}]{Liu+19}
Yinhan Liu, Myle Ott, Naman Goyal, Jingfei Du, Mandar Joshi, Danqi Chen, Omer
  Levy, Mike Lewis, Luke Zettlemoyer, and Veselin Stoyanov. 2019.
\newblock \href {https://arxiv.org/abs/1907.11692} {{RoBERTa}: {A} robustly
  optimized {BERT} pretraining approach}.
\newblock \emph{arXiv preprint arXiv:1907.11692}, pages 1--13.

\bibitem[{Pires et~al.(2020)Pires, Schlinger, and Garrette}]{Pires+20}
Telmo Pires, Eva Schlinger, and Dan Garrette. 2020.
\newblock \href {https://arxiv.org/abs/1906.01502} {How multilingual is
  multilingual {BERT}?}
\newblock \emph{arXiv preprint arXiv:1906.01502}, pages 1--6.

\bibitem[{Rajpurkar et~al.(2016)Rajpurkar, Zhang, Lopyrev, and
  Liang}]{Rajpurkar+16}
Pranav Rajpurkar, Jian Zhang, Konstantin Lopyrev, and Percy Liang. 2016.
\newblock {SQuAD}: 100,000+ questions for machine comprehension of text.
\newblock In \emph{Proceedings of the Conference on Empirical Methods in
  Natural Language Processing (EMNLP)}, pages 2383--2392, Austin, Texas.

\bibitem[{Tiedemann(2012)}]{Tiedemann-12}
J{\"o}rg Tiedemann. 2012.
\newblock Parallel data, tools and interfaces in {OPUS}.
\newblock In \emph{Proceedings of the Eighth International Conference on
  Language Resources and Evaluation (LREC)}, pages 2214--2218, Istanbul.

\bibitem[{van~den Oord et~al.(2019)van~den Oord, Li, and Vinyals}]{Oord+19}
Aaron van~den Oord, Yazhe Li, and Oriol Vinyals. 2019.
\newblock \href {https://arxiv.org/abs/1807.03748} {Representation learning
  with contrastive predictive coding}.
\newblock \emph{arXiv preprint arXiv:1807.03748}, pages 1--13.

\bibitem[{Vaswani et~al.(2017)Vaswani, Shazeer, Parmar, Uszkoreit, Jones,
  Gomez, Kaiser, and Polosukhin}]{Vaswani+17}
Ashish Vaswani, Noam Shazeer, Niki Parmar, Jakob Uszkoreit, Llion Jones,
  Aidan~N. Gomez, {\L}ukasz Kaiser, and Illia Polosukhin. 2017.
\newblock Attention is all you need.
\newblock In \emph{Advances in Neural Information Processing Systems (NIPS)},
  pages 5998--6008, Long Beach, CA.

\bibitem[{Wang et~al.(2019)Wang, Che, Guo, Liu, and Liu}]{Wang+19}
Yuxuan Wang, Wanxiang Che, Jiang Guo, Yijia Liu, and Ting Liu. 2019.
\newblock Cross-lingual {BERT} transformation for zero-shot dependency parsing.
\newblock In \emph{Proceedings of the Conference on Empirical Methods in
  Natural Language Processing and the International Joint Conference on Natural
  Language Processing (EMNLP-IJCNLP)}, pages 5721--5727, Hong Kong.

\bibitem[{Wei et~al.(2020)Wei, Hu, Weng, Xing, Yu, and Luo}]{Wei+20}
Xiangpeng Wei, Yue Hu, Rongxiang Weng, Luxi Xing, Heng Yu, and Weihua Luo.
  2020.
\newblock \href {https://arxiv.org/abs/2007.15960} {On learning universal
  representations across languages}.
\newblock \emph{arXiv preprint arXiv:2007.15960}, pages 1--13.

\bibitem[{Williams et~al.(2018)Williams, Nangia, and Bowman}]{Williams+18}
Adina Williams, Nikita Nangia, and Samuel Bowman. 2018.
\newblock A broad-coverage challenge corpus for sentence understanding through
  inference.
\newblock In \emph{Proceedings of the 19th Annual Conference of the North
  American Chapter of the Association for Computational Linguistics: Human
  Language Technologies (HLT-NAACL)}, pages 1112--1122, New Orleans. The
  Association for Computational Linguistics.

\bibitem[{Wu and Dredze(2019)}]{Wu-Dredze-19}
Shijie Wu and Mark Dredze. 2019.
\newblock {B}eto, {B}entz, {B}ecas: The surprising cross-lingual effectiveness
  of {BERT}.
\newblock In \emph{Proceedings of the Conference on Empirical Methods in
  Natural Language Processing and the International Joint Conference on Natural
  Language Processing (EMNLP-IJCNLP)}, pages 833--844, Hong Kong.

\bibitem[{Yang et~al.(2020)Yang, Ma, Zhang, Wu, Li, and Zhou}]{Yang+20}
Jian Yang, Shuming Ma, Dongdong Zhang, ShuangZhi Wu, Zhoujun Li, and Ming Zhou.
  2020.
\newblock Alternating language modeling for cross-lingual pre-training.
\newblock In \emph{Proceedings of the 34th Conference on Artificial
  Intelligence (AAAI)}, pages 9386--9393, New York. AAAI Press.

\bibitem[{Ziemski et~al.(2016)Ziemski, Junczys-Dowmunt, and
  Pouliquen}]{Ziemski+16}
Micha{\l} Ziemski, Marcin Junczys-Dowmunt, and Bruno Pouliquen. 2016.
\newblock The {U}nited {N}ations parallel corpus v1.0.
\newblock In \emph{Proceedings of the Tenth International Conference on
  Language Resources and Evaluation (LREC)}, pages 3530--3534, Portoro{\v{z}},
  Slovenia.

\end{thebibliography}
\bibliographystyle{acl_natbib}

\end{document}